\pdfoutput=1

\documentclass[11pt]{article}

\usepackage{ACL2023}
\usepackage{graphicx}
\usepackage{subcaption}
\usepackage{booktabs}
\usepackage{multirow}
\usepackage{tabularx}
\usepackage{xcolor}
\usepackage{caption}
\usepackage{listings}
\usepackage{times}
\usepackage{latexsym}

\usepackage[T1]{fontenc}

\usepackage[utf8]{inputenc}

\usepackage{microtype}

\usepackage{inconsolata}

\usepackage{algorithm}
\usepackage{algorithmic}

\title{Enhancing Text Classification through LLM-Driven Active Learning and Human Annotation}

\author{Hamidreza Rouzegar \and Masoud Makrehchi \\[2ex] 
\textbf{Department of Electrical, Computer and Software Engineering} \\ 
\textbf{Ontario Tech University} \\ 
Oshawa, ON, Canada \\ 
\texttt{hamidreza.rouzegar@ontariotechu.net} \\ 
\texttt{masoud.makrehchi@ontariotechu.ca}\\ 
\\
}

\begin{document}
\maketitle

\begin{abstract}
In the context of text classification, the financial burden of annotation exercises for creating training data is a critical issue. Active learning techniques, particularly those rooted in uncertainty sampling, offer a cost-effective solution by pinpointing the most instructive samples for manual annotation. Similarly, Large Language Models (LLMs) such as GPT-3.5 provide an alternative for automated annotation but come with concerns regarding their reliability. This study introduces a novel methodology that integrates human annotators and LLMs within an Active Learning framework. We conducted evaluations on three public datasets. IMDB for sentiment analysis, a Fake News dataset for authenticity discernment, and a Movie Genres dataset for multi-label classification.
The proposed framework integrates human annotation with the output of LLMs, depending on the model uncertainty levels. This strategy achieves an optimal balance between cost efficiency and classification performance. The empirical results show a substantial decrease in the costs associated with data annotation while either maintaining or improving model accuracy.
\end{abstract}

\section{Introduction}

Active learning allows machine learning algorithms to choose their learning data selectively. This methodology optimizes learning efficiency while reducing the need for labor-intensive labeled instances, often leading to enhancing the performance metrics with less training \citep{settles2008active}.

While Active Learning has been explored in research for over two decades, it has seen a resurgence of interest in Natural Language Processing (NLP), particularly around 2009-2010. This resurgence has been aligned with adopting neural models in NLP research \citep{zhang-etal-2022-survey}. Recent trends suggest the advantageous pairing of Active Learning techniques with deep learning methodologies \citep{zhang-etal-2022-survey}.

A prevalent strategy within Active Learning is uncertainty sampling. This technique involves the algorithm selecting instances in which the model might be least certain about their labels. For instance, in binary classification tasks, it might select an instance whose probability of belonging to the positive class is nearest to 0.5. This often employs a 'pool-based' approach where a human expert validates and assigns true labels to the selected samples, which are then used to update the classifier iteratively \citep{Lewis-etal-1994}. Additional approaches like Query-By-Committee (QBC) \citep{seung1992query} and Expected Gradient Length (EGL) \citep{settles2009active} offer alternative techniques in Active Learning.

Broadly speaking, Active Learning aims to reduce the expenses associated with human annotation, achieving this by strategically selecting the most informative data points for labeling \citep{hachey-etal-2005-investigating}. However, Active Learning is not the only strategy employed to minimize human annotation costs. LLMs have been applied to various text annotation tasks such as political Twitter message categorization \citep{tornberg2023chatgpt}, relevance and topic detection in tweets \citep{gilardi2023chatgpt}, and hate speech classification \citep{huang2023chatgpt}. Some recent studies have also examined the potential of utilizing Active Learning in Prompt-Based Uncertainty sampling with LLMs \citep{yu-etal-2023-cold}.

In this study, we propose a novel pipeline for text classification focusing on three distinct, open-source datasets: IMDB for sentiment analysis, a dataset for identifying fake news, and another for classifying Movie Genres. We introduce a framework that integrates Active Learning based on uncertainty sampling with human and GPT-3.5 annotations. This integration is tailored to adaptively choose between human and machine annotators based on the uncertainty levels estimated from GPT-3.5's annotation.

To the best of our knowledge, this is the first study that comprehensively evaluates the utility and efficiency of combining human annotators, Active Learning, and GPT-3.5 in a text classification task. We extend the traditional Active Learning methodologies by integrating uncertainty measurements from LLM, such as GPT-3.5, into our annotation selection process. This not only minimizes the costs of manual annotation but also capitalizes on the strengths of machine learning models for efficient and accurate text classification.

Our approach offers a nuanced trade-off analysis between cost and accuracy, using real-world pricing models for both human and machine annotation. The goal is to provide a robust, cost-effective, and scalable text classification pipeline that leverages the best of both human expertise and advanced machine learning techniques. The code provide in anonymous GitHub.\footnote{\href{https://github.com/hrouzegar/Enhancing-Text-Classification-through-LLM-Driven-Active-Learning-and-Human-Annotation.git}{GitHub Code}}

In the subsequent sections, we explore a methodology that combines human expertise with LLMs in an Active Learning framework. We detail the experimental setup and methodologies, including uncertainty-based sampling and LLMs for data annotation. The results are analyzed for their accuracy and efficiency, leading to a discussion on the broader implications and future potential of this approach in text classification.

\section{Related Works}

Text classification is an instrumental task in Natural Language Processing (NLP) that uses methods ranging from traditional machine learning techniques to advanced neural networks such as  Long Short-Term Memory (LSTM)  \citep{qaisar2020sentiment} and Convolutional Neural Networks (CNN) \citep{haque2019performance}. Although these methods primarily focus on improving model accuracy, our study diverges by emphasizing the reduction of labeled data through selective annotation, subsequently enhancing the model performance. We also manage the pool of unlabeled data by removing instances that have been labeled, thus continuously refining our dataset for model training.

\subsection{Uncertainty-based Active Learning}
Anderson et al. \citep{andersen2022towards} explored Active Learning based on uncertainty across different models, including support vector machines, logistic regression, and decision trees. They proposed a criterion for manual annotation that identifies a certain percentage of the most uncertain predictions for each model type as a stopping point for annotation. For example, they suggested a 12.71\% threshold for logistic regression to achieve desired model performance.

In \citep{goudjil2018novel} Active Learning is applied on text analysis tasks. The authors used support vector machines and Active Learning-supported SVM (AL-SVM) models. They introduced a thresholding mechanism, such as setting a 70\% threshold for AL-SVM, to select instances with less than 70\% probability for annotation. 

\subsection{Advanced Active Learning Methods}
In \citep{yuan-etal-2020-cold}, a range of Active Learning techniques, such as Active Learning by Processing Surprisal and Entropy, are examined. However, their work did not establish any clear stopping criterion to determine how many Active Learning iterations are necessary. 

In the work by Zhang et al. \citep{zhang2022allsh}, they proposed a method that uses local sensitivity and adversarial perturbations in the Active Learning loop. They aimed to alleviate the sampling bias that comes from selecting only the most uncertain examples. They used data augmentation and adversarial perturbation to measure the local sensitivity of instances, thereby picking examples that lie near the decision boundary. Their approach also incorporated a "learning hardness" criterion to sidestep examples that are hard to learn or potentially mislabeled. 

\subsection{Active Learning with Language Models}
Yue Yu et al. \citep{yu-etal-2023-cold} proposed a method called PATRON that integrates prompt-based techniques for cold-start data selection in Active Learning. Their method uses estimated uncertainty for data points, adopting two key design strategies—uncertainty propagation and a partition-then-rewrite (PTR) strategy to ensure a balance between informativeness and diversity in sample selection.

Building upon these foundations, recent studies have furthered our understanding of the potential of Active Learning when combined with LLMs. Zhang et al. \citep{zhang2023llmaaa} introduced a framework where LLMs serve as active annotators, underscoring the efficiency of LLMs in reducing annotation costs while maintaining accuracy. "Beyond Labels" further expanded this concept by integrating human annotators with machine-generated natural language explanations, demonstrating the potential for more informative annotations in low-resource settings.

In \citep{xiao2023freeal}, the authors proposed a novel collaborative learning approach, reducing the need for human annotation by leveraging LLMs as weak annotators. This framework indicates the potential of LLMs to enhance unsupervised performance in various NLP tasks. On the other side, Lu et al.  \citep{lu2023human} presented empirical evidence that smaller models trained with expert annotations can outperform LLMs in domain-specific tasks, highlighting the irreplaceable value of human expertise.

Furthermore, Margatina et al. \citep{margatina2023active} explored selecting demonstrations for few-shot learning with LLMs, revealing that demonstrations semantically similar to test examples yield superior performance across tasks and model sizes. This finding supports our methodology of integrating human annotators and GPT-3.5 in a new Active Learning paradigm evaluated across multiple open-source datasets.

Lastly, in \citep{margatina2023limitations}, the authors critically examined the challenges in Active Learning research, especially in simulated environments. They emphasize the need for realistic, transparent, and reproducible Active Learning research, aligning with our approach of combining human expertise with the capabilities of LLMs to address the challenges in text classification and Active Learning.

While these works have significantly advanced the field of Active Learning, there remains a gap in leveraging the power of LLMs like GPT-3.5 for both annotation and uncertainty estimation in an integrated Active Learning framework. Our work aims to bridge this gap by using both human annotators and GPT-3.5 in a novel Active Learning paradigm, which we evaluate across multiple open-source datasets.

\begin{figure*}[t]

   \centering
    \captionsetup{justification=raggedright,singlelinecheck=false}
   \includegraphics[width=\textwidth]{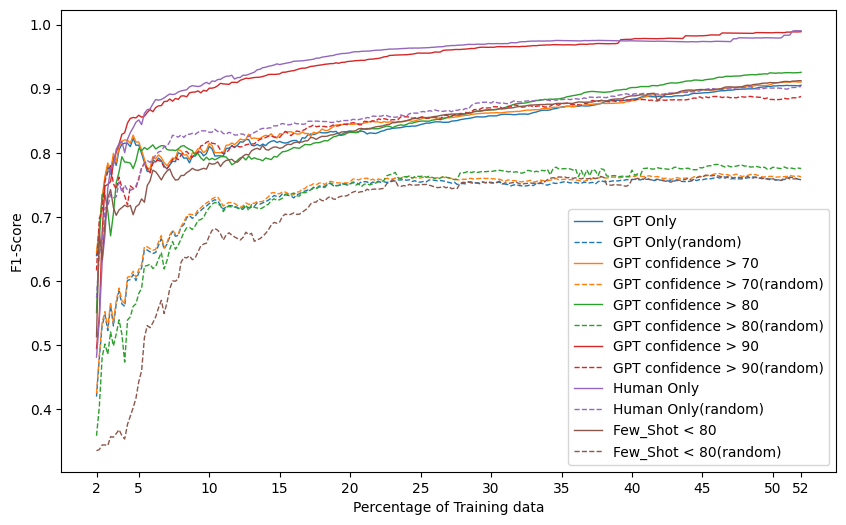}
   \caption{\label{fig:NEWS}
F1 Score Progression from 2\% to 52\% Training Data Portions in the Fake News Dataset. This figure visualizes the evolution of F1 scores across different training data portions, ranging from 2\% to 52\%, for various annotation methods, including GPT-only, Hybrid models, Human-only, Few-shot learning, and a baseline of random sampling. Each incremental step represents an increase in the training dataset size, highlighting the performance changes in F1 scores across the experiments.}
\end{figure*}
\begin{figure*}[t] 

   \centering
   \captionsetup{justification=raggedright,singlelinecheck=false}
   \includegraphics[width=\textwidth]{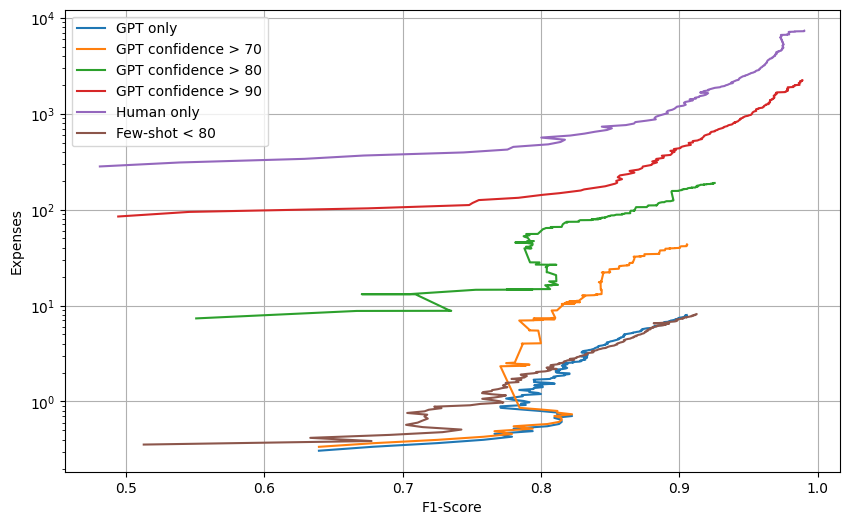}

   \caption{\label{fig:News_cost}
Cost per F1 Score Analysis During Iterative Training Data Increments in the Fake News Dataset. This figure illustrates the cost efficiency (cost per F1 score) for different annotation strategies as the training data portion increases from 2\% to 52\%. It compares the cost-effectiveness of GPT-only, Hybrid models, Human-only, and Few-shot learning methods, providing insights into the financial implications of each annotation approach over the iterative training process.}
\end{figure*}

\section{Methodology}
This section details the methodology employed for text classification, emphasizing the integration of Large Language Models in data annotation. The approach encompasses an Active Learning framework, applying uncertainty-based sampling across varied datasets to enhance accuracy and efficiency.
\begin{algorithm}
\caption{Procedure for Collecting Sentiment Label and Confidence}
\label{Sentiment}
\begin{algorithmic}[1]
\REQUIRE Movie review text $X[i]$
\ENSURE Sentiment label and confidence level in JSON format

\STATE \textbf{Prompt:} ``What is the sentiment of the following movie review, and how confident are you about this 'sentiment'?"
\STATE \textbf{Instructions:} ``Give your answer as a single word, either 'positive' or 'negative' and a single percentage in JSON format delimited with space."
\STATE \textbf{Display:} ``Review text: ''' $X[i]$ '''"

\end{algorithmic}
\end{algorithm}

\subsection{Active Learning based on uncertainty sampling}
Our Active Learning approach revolved around the concept of uncertainty sampling. In each iteration of the Active Learning process, we used an iterative strategy to select the most uncertain data points from the unlabeled data pool, employing the logistic regression model's predicted probabilities. The ranking of data points based on their predicted probabilities provided a measure of uncertainty or confidence measure for each sample. This ranking helps to identify and select the data points for which the classifier was most uncertain, enabling the model to learn from these challenging instances and refine its classification performance.

\textbf{Data Selection:}
In each Active Learning iteration,  the data samples with the highest uncertainty scores from the pool of unlabeled data are selected, adding these data samples to the training set for the next iteration. This process of focusing on high-information gain data allowed the model to learn from its previous mistakes and incrementally improve its overall performance with each iteration. By iteratively selecting and incorporating informative data, the Active Learning approach optimizes the learning process efficiently, requiring significantly fewer labeled samples compared to conventional supervised learning methods.

\textbf{Pool Initiation:}
We initiated the Active Learning process by creating an initial set of data that incorporated a small fraction of the entire data available for the classification task. This set served as the starting point for the learning process, enabling the model to make its initial predictions. The rest of the data, not included in the initial set, constituted the 'unlabeled pool.' This pool continuously provided a source of uncertain samples to be selected and labeled during each iteration of the Active Learning process.

\subsection{Proxy-Validation Set}
One of the key contributions of our work is the creation of a 'proxy-validation' set. This set, which is a subset of the total data, served to estimate the model's performance at each iteration of the Active Learning process, acting as a set of labeled samples. It also emulated the statistical distribution presented in the main unlabeled pool, undergoing updates alongside each iteration.

During each Active Learning iteration, we computed the model's accuracy on the proxy-validation set. To ensure consistency, we applied the same percentage of confidence for low-confidence data removal to the proxy-validation set as we did to the main unlabeled pool. The remaining data in the proxy-validation set provided us with an estimation of the main pool's accuracy, a crucial measure when true labels for the pool were unavailable.

\subsection{LLM-based Data Annotation}
We employed the GPT-3.5 API to annotate our dataset, increasing the overall efficiency of the Active Learning process. A set of prompts is designed for GPT-3.5 to predict the sentiment of movie reviews and to report confidence in each prediction. The use of LLM for annotation allowed us to obtain sentiment labels and corresponding confidence scores for each data, opening the way for several experimental conditions. The procedure for IMDB datasets is illustrated in Algorithm \ref{Sentiment}.

\textbf{Adaptation of Active Learning to Prompt Engineering:}
Building on the foundational concepts of Active Learning,  a new approach is introduced to apply the components mentioned above in a solution based on  LLMs such as GPT-3.5.

Unlike usual Active Learning methods, which often require retraining models with carefully chosen labeled data, our method takes a different path by focusing on improving the prompts given to GPT-3.5. This approach makes the most of the model's pre-existing knowledge and avoids the need for retraining. Instead, GPT-3.5's responses carefully crafted the prompts.

Initially, we utilize zero-shot learning, presenting tasks to GPT-3.5 without any specific examples. The model’s response and associated confidence scores provide an initial measure of its proficiency in the given task. These confidence scores are analogous to uncertainty measures in traditional Active Learning, guiding our subsequent steps.

In instances where GPT-3.5 exhibits lower confidence (below 70\% in the IMDB dataset and below 80\% in two other datasets), we transition to a few-shot learning approach. This progression involves providing the model with low-confidence tasks. These thresholds were chosen based on an analysis of our datasets, which revealed that lower thresholds did not significantly change many labels, while higher thresholds led to an excessively large portion of the data being re-annotated. With these thresholds, we managed to target approximately 10 to 15 percent of the IMDB and Movie Genres datasets and about 4 percent of the fake news dataset, ensuring a manageable yet effective scope for applying few-shot learning.

This methodology effectively replicates the essence of Active Learning, where the model iteratively improves by focusing on the most informative or uncertain samples. By applying this approach to prompt design, we leverage the innate capabilities of LLMs for more efficient and targeted learning.

\section{Experimental Setup}

\begin{table*}[h]
\centering
\captionsetup{justification=raggedright,singlelinecheck=false} 
{\footnotesize
\begin{tabularx}{\textwidth}{@{}l|>{\centering\arraybackslash}X|>{\centering\arraybackslash}X|>{\centering\arraybackslash}X|>{\centering\arraybackslash}X|>{\centering\arraybackslash}X|>{\centering\arraybackslash}X|>{\centering\arraybackslash}X|>{\centering\arraybackslash}X|>{\centering\arraybackslash}X|>{\centering\arraybackslash}X@{}}
\toprule
\multicolumn{1}{c|}{\textbf{Portion}} & \multicolumn{2}{c}{\textbf{10\%}} & \multicolumn{2}{c}{\textbf{20\%}} & \multicolumn{2}{c}{\textbf{30\%}} & \multicolumn{2}{c}{\textbf{40\%}} & \multicolumn{2}{c}{\textbf{50\%}} \\ \cmidrule(l){2-11} 
\multicolumn{1}{c|}{\textbf{Method}} & Cost & F1 & Cost & F1 & Cost & F1 & Cost & F1 & Cost & F1 \\ \midrule
\textbf{GPT only} & 0.46 & 0.8201 & 0.92 & 0.8651 & 1.3809 & 0.9152 & 1.84 & 0.9439 & 2.30 & 0.9629 \\ \midrule
\textbf{GPT conf > 70} & 74.44 & 0.8548 & 170.45 & 0.9077 & 283.37 & 0.9446 & 373.04 & 0.9671 & 442.42 & 0.978 \\ \midrule
\textbf{GPT conf > 80} & 219.04 & 0.8522 & 476.13 & 0.908 & 733.22 & 0.9469 & 958.60 & 0.9696 & 1151.85 & 0.9802 \\ \midrule
\textbf{GPT conf > 90} & 369.12 & 0.8533 & 750.09 & 0.9112 & 1139.51 & 0.9495 & 1517.10 & 0.9700 & 1873.55 & 0.9800 \\ \midrule
\textbf{Human only} & 423.24 & 0.8597 & 846.49 & 0.9085 & 1269.73 & 0.9475 & 1692.98 & 0.9693 & 2116.22 & 0.9796 \\ \midrule
\textbf{Few-shot <70} & 0.95 & 0.8469 & 1.77 & 0.8973 & 2.60 & 0.9407 & 3.33 & 0.9631 & 4.07 & 0.9773 \\ \bottomrule
\end{tabularx}
}
\caption{\label{tab:imdb_dataset_results}
IMDB Dataset Results: F1 Scores and Costs (in USD) for Various Annotation Methods on the IMDB Dataset. This table illustrates the comparative performance and cost-efficiency of different annotation approaches, including GPT-only, Hybrid models with varying confidence thresholds, Human-only, and Few-shot learning strategies across 10\% to 50\% data portions.}

\end{table*}

\begin{table*}[h]
\centering
\captionsetup{justification=raggedright,singlelinecheck=false} 
{\footnotesize
\begin{tabularx}{\textwidth}{@{}l|>{\centering\arraybackslash}X|>{\centering\arraybackslash}X|>{\centering\arraybackslash}X|>{\centering\arraybackslash}X|>{\centering\arraybackslash}X|>{\centering\arraybackslash}X|>{\centering\arraybackslash}X|>{\centering\arraybackslash}X|>{\centering\arraybackslash}X|>{\centering\arraybackslash}X@{}}
\toprule
\multicolumn{1}{c|}{\textbf{Portion}} & \multicolumn{2}{c}{\textbf{10\%}} & \multicolumn{2}{c}{\textbf{20\%}} & \multicolumn{2}{c}{\textbf{30\%}} & \multicolumn{2}{c}{\textbf{40\%}} & \multicolumn{2}{c}{\textbf{50\%}} \\ \cmidrule(l){2-11} 
\multicolumn{1}{c|}{\textbf{Method}} & Cost & F1 & Cost & F1 & Cost & F1 & Cost & F1 & Cost & F1 \\ \midrule
\textbf{GPT only} & 1.53 & 0.8099 & 3.06 & 0.8339 & 4.59 & 0.8576 & 6.12 & 0.8837 & 7.66 & 0.9041 \\ \midrule
\textbf{GPT conf > 70} & 7.16 & 0.8099 & 12.91 & 0.8394 & 25.70 & 0.8573 & 34.27 & 0.8764 & 41.44 & 0.9028 \\ \midrule
\textbf{GPT conf > 80} & 39.53 & 0.7889 & 77.66 & 0.8319 & 97.49 & 0.8675 & 159.54 & 0.9017 & 187.82 & 0.9243 \\ \midrule
\textbf{GPT conf > 90} & 437.85 & 0.9007 & 843.33 & 0.9426 & 1289.63 & 0.9646 & 1714.82 & 0.9771 & 2156.90 & 0.9871 \\ \midrule
\textbf{Human only} & 1409.02 & 0.9074 & 2818.04 & 0.956 & 4227.07 & 0.9701 & 5636.09 & 0.9745 & 7045.11 & 0.9791 \\ \midrule
\textbf{Few-shot <70} & 1.59 & 0.7838 & 3.14 & 0.8335 & 4.69 & 0.8667 & 6.24 & 0.8888 & 7.81 & 0.9092 \\ \bottomrule
\end{tabularx}
}
\caption{\label{tab:fake_news_dataset_results}
Fake News Dataset Results: Comparative Analysis of F1 Scores and Annotation Costs (in USD) for the Fake News Dataset. The table presents a detailed breakdown of the performance metrics and financial implications of using GPT-only, various Hybrid models, Human-only, and Few-shot annotations at different data portions.}
\end{table*}

\begin{table*}[h]
\centering
\captionsetup{justification=raggedright,singlelinecheck=false} 
{\footnotesize
\begin{tabularx}{\textwidth}{@{}l|>{\centering\arraybackslash}X|>{\centering\arraybackslash}X|>{\centering\arraybackslash}X|>{\centering\arraybackslash}X|>{\centering\arraybackslash}X|>{\centering\arraybackslash}X|>{\centering\arraybackslash}X|>{\centering\arraybackslash}X|>{\centering\arraybackslash}X|>{\centering\arraybackslash}X@{}}
\toprule
\multicolumn{1}{c|}{\textbf{Portion}} & \multicolumn{2}{c}{\textbf{10\%}} & \multicolumn{2}{c}{\textbf{20\%}} & \multicolumn{2}{c}{\textbf{30\%}} & \multicolumn{2}{c}{\textbf{40\%}} & \multicolumn{2}{c}{\textbf{50\%}} \\ \cmidrule(l){2-11} 
\multicolumn{1}{c|}{\textbf{Method}} & Cost & F1 & Cost & F1 & Cost & F1 & Cost & F1 & Cost & F1 \\ \midrule
\textbf{GPT only} & 0.81 & 0.3530 & 1.62 & 0.4066 & 2.43 & 0.4236 & 3.24 & 0.4233 & 4.05 & 0.4337 \\ \midrule
\textbf{GPT conf > 70} & 6.01 & 0.3767 & 8.31 & 0.4056 & 17.31 & 0.4235 & 25.56 & 0.4308 & 31.58 & 0.4364 \\ \midrule
\textbf{GPT conf > 80} & 125.82 & 0.2688 & 169.79 & 0.4806 & 233.10 & 0.4998 & 299.39 & 0.5053 & 347.83 & 0.5358 \\ \midrule
\textbf{GPT conf > 90} & 263.48 & 0.6212 & 534.40 & 0.716 & 809.79 & 0.7513 & 1078.48 & 0.7959 & 1353.87 & 0.8271 \\ \midrule
\textbf{Human only} & 744.92 & 0.6039 & 1489.84 & 0.711 & 2234.77 & 0.7520 & 2979.69 & 0.8099 & 3724.61 & 0.8443 \\ \midrule
\textbf{Few-shot <70} & 0.88 & 0.37 & 1.74 & 0.4223 & 2.60 & 0.4726 & 3.4615 & 0.4761 & 4.31 & 0.4996 \\ \bottomrule
\end{tabularx}
}
\caption{\label{tab:movie_genre_dataset_results}
Movie Genres Dataset Results: Performance Metrics and Costs (in USD) Across Different Annotation Methods for the Movie Genres Dataset. This table compares the F1 scores and associated costs for GPT-only, Hybrid annotation at different confidence levels, Human-only, and Few-shot learning across incremental portions of the dataset.}
\end{table*}

In the experimental setup, we explore the impact of using GPT-3.5 for data annotation under various scenarios, including different confidence thresholds and combinations of human and GPT-3.5 annotations. The experiments were conducted across three datasets: IMDB, Movie Genress, and fake news.

\textbf{GPT-3.5 Labels Only}
Our first experiment investigates the feasibility of employing an LLM for data annotation using only the labels provided by GPT-3.5.

\textbf{Human Labels Only}
As a baseline, our second experiment involves using human annotations exclusively. This experiment serves as a control to measure the effectiveness of GPT-3.5’s annotations against traditional human annotation.

\textbf{Hybrid Labels: Confidence Threshold Experiments}
These experiments explore the efficacy of combining GPT-3.5's predictions with human annotations at various confidence levels set by GPT-3.5.

\textbf{Confidence Threshold 90}
For data points where GPT-3.5's confidence score exceeds 90\%, we utilize the labels provided by the model. If the confidence score is below 90\%, human annotations are used. This experiment aims to evaluate the performance and cost-effectiveness of relying predominantly on AI annotations at a high confidence level.

\textbf{Confidence Threshold 80}
In this setup, GPT-3.5's labels are adopted for data points with a confidence score above 80\%. For those below this threshold, human annotations are employed. This approach aims to balance AI efficiency and human accuracy at an intermediate confidence level.

\textbf{Confidence Threshold 70}
Here, the threshold is set at 70\% confidence. GPT-3.5's labels are used for data points above this level, while human annotations supplement the lower-confidence points. The objective is to assess the impacts of a lower threshold on annotation efficiency and accuracy.

\textbf{GPT-3.5 with Few-Shot Learning for Active Learning}
This experiment investigates the application of GPT-3.5's few-shot learning for data annotation in an Active Learning context, specifically focusing on data points with varying confidence levels. Unlike the hybrid approach that combines human and GPT-3.5 annotations, this setup utilizes GPT-3.5's few-shot learning capabilities exclusively.
The aim here is to assess how GPT-3.5's few-shot learning can enhance the model's annotation performance, particularly for data points where it initially shows low confidence. The strategy involves:

{One-Shot Learning for Higher Confidence Data Points:}
For data points where GPT-3.5's confidence score is above a certain threshold, we employ one-shot learning. GPT-3.5 is provided with a single relevant example to refine its understanding and improve annotations.

{Few-Shot Learning for Lower Confidence Data Points:} For data points with confidence scores below the threshold, few-shot learning is implemented, where GPT-3.5 is given three examples to assist its annotations.

\textbf{Confidence Thresholds:}
In the IMDB Dataset, A 70\% confidence threshold is used. Data points above this threshold receive one-shot learning, while those below it are processed with few-shot learning.
However, for Movie Genres and Fake News Datasets, An 80\% confidence threshold is applied. Similarly, data points above this threshold are handled with one-shot learning and those below it with few-shot learning.

\textbf{Baseline Comparison (Random Data Addition): }
Each experimental setup includes a component where data points are added to the training set randomly, serving as a baseline for comparison. This strategy illustrates the advantages of our more targeted Active Learning approaches.

{Comparison and Cost Estimation:}
We evaluate each experimental setup based on two main metrics: the F1 score and the associated annotation costs. The focus is on finding the optimal balance between accuracy and cost-efficiency, particularly in the hybrid annotation and few-shot learning scenarios. The results highlight the trade-offs involved in using GPT-3.5's annotations to reduce costs.

\textbf{Cost Estimation:}
For a more thorough understanding of the feasibility of each approach, the annotation costs associated with each experiment were evaluated. These costs were computed based on the pricing structure provided by the AI Platform Data Labeling Service for human labels\footnote{\url{https://cloud.google.com/ai-platform/data-labeling/pricing}}, and the pricing for the GPT-3.5 API \footnote{\url{https://openai.com/pricing}} for LLM generated labels.
For human annotation costs, we referred to the AI Platform Data Labeling Service pricing, which uses units per human labeler, with each unit encompassing a fixed number of words. The total cost of human annotation was calculated by multiplying the total number of units labeled by human annotators with the per-unit price.
For LLM annotation costs, we utilized the pricing structure provided for the GPT-3.5 API, which charges based on the number of tokens processed. The total cost of machine annotation was determined by multiplying the total number of tokens processed by GPT-3.5 with the cost per token.
By taking both F1-Score and cost into consideration, we established a comprehensive comparison of each experiment. This comparison provided insights into the trade-off between cost and accuracy, allowing us to identify the most economically efficient approach that does not compromise the performance of our Active Learning model. This comprehensive evaluation and comparison serve as a valuable guide in implementing Active Learning strategies for text classification tasks.

\section{Analysis and Results}
The methodology exhibited notable scalability across three distinct datasets, each presenting unique classification challenges:
IMDB Reviews: This dataset involved a binary classification task, determining the sentiment of movie reviews as either positive or negative.
Fake News: Another binary classification task where the variability in text lengths presented an additional layer of complexity.
Movie Genres: This dataset represented a more intricate four-class classification based on movie plots.
The consistent application of the methodology across these datasets underscores its adaptability and versatility in handling varying text lengths and classification complexities.

The F1 scores and costs associated with each experimental setup were analyzed. For instance, in the IMDB dataset, GPT-only annotations demonstrated a progressive increase in F1 scores from 0.8201 at 10\% to 0.9629 at 50\%, with corresponding costs ranging from 0.4603 to 2.3015. Conversely, the human-only approach showed F1 scores from 0.8597 to 0.9796, with significantly higher costs.(Table \ref{tab:imdb_dataset_results})

Critically, Figure \ref{fig:NEWS} demonstrates that all annotation methods significantly outperform random sampling. This superiority is evident across various metrics, particularly as the training data incrementally increases from 2\% to 52\%. These findings underscore the effectiveness of structured annotation strategies over random approaches in enhancing the efficiency and accuracy of text classification.

Confidence thresholds played a crucial role in balancing automated and manual annotations. In the IMDB dataset, a 70\% threshold was used, where data points above this threshold received one-shot learning, and those below it were processed with few-shot learning. For the Movie Genres and Fake News datasets, an 80\% threshold was applied. These thresholds were determined based on the datasets' characteristics, targeting approximately 10 to 15 percent of the IMDB and Movie Genres datasets and about 4 percent of the Fake News dataset for few-shot learning. This strategy effectively replicated the essence of Active Learning, focusing on the most informative or uncertain samples.

The cost implications of each setup were a focal point. For instance, the Hybrid 80 model for IMDB and the Hybrid 90 for Movie Genres demonstrated significant cost-efficiency while achieving comparable accuracies to human-only labels but at a fraction of the cost. This was effectively illustrated through logarithmic scale representations, Figure \ref{fig:News_cost} highlighting stark cost disparities and indicating the practicality of the chosen thresholds and annotation strategies.

The concept of proxy validation emerged as a crucial aspect of the study. Analysis revealed a notable correlation between the F1 score of the proxy validation and the remaining pool, indicating that proxy validation serves as a reliable indicator of the overall pool quality. Though some variations between the F1 score of proxy validation and the actual pool were observed, these discrepancies were minimal. Detailed examples and a deeper analysis of this phenomenon are provided in the appendix. Increasing the size of the proxy validation compared to pool data might lead to even more similar F1 scores between proxy validation and the pool, enhancing the reliability of this method as an indicator.

The study also delved into the analysis of GPT-3.5's output confidence scores. In the IMDB dataset, 11\% of the annotations were found to be incorrect overall. However, in instances where GPT-3.5's confidence was below 70\%, the rate of incorrect annotations rose to nearly 50\%. Similar trends were observed in the Movie Genres dataset (33\% overall inaccuracies, rising to 50\% for data under 80\% confidence) and the Fake News dataset (27\% overall inaccuracies, increasing to nearly 50\% for data under 80\% confidence). These findings suggest that the model's confidence score can be a reliable indicator of uncertainty and the likelihood of annotation errors.

These results have significant implications for the application of Active Learning models, particularly in how confidence scores can be interpreted and used. The observed correlation between lower confidence scores and higher rates of annotation errors supports the idea that GPT-3.5's confidence can be treated similarly to uncertainty measures in traditional Active Learning models. This insight opens up new avenues for utilizing LLMs in Active Learning frameworks, where confidence scores can guide the annotation process more effectively.

\section*{Conclusion}
The paper demonstrates that combining Large Language Models (LLMs), such as GPT-3.5, with human annotators in an Active Learning framework significantly enhances text classification tasks. This hybrid approach, which selectively employs either GPT-3.5 or human annotations based on confidence thresholds, efficiently balances cost and accuracy. The methodology reduces data annotation expenses while maintaining or even improving model performance compared to traditional human-only annotation methods. The study also introduces the concept of proxy validation, which effectively estimates the quality of the entire unlabeled dataset, proving useful in optimizing the annotation process. Overall, the research highlights the benefits of integrating advanced AI models with human insights to create more efficient, accurate, and scalable solutions for text classification.
\bibliography{anthology,custom}
\bibliographystyle{acl_natbib}

\appendix

\clearpage

\section{Appendix}
\label{sec:appendix}

In this appendix, additional figures and detailed explanations to supplement the findings and methodologies presented in the main body of the study are provided. These supplementary materials are crucial for a deeper understanding of the performance and efficiency of our proposed Active Learning framework, as well as the effectiveness of our proxy validation approach and prompt design strategy in handling various text classification tasks. Each section of the appendix is dedicated to a specific aspect of our research, offering visual representations, example prompts, and thorough descriptions of the datasets and Active Learning process used in our study.
\subsection{Additional Figures of Active Learning Performance and Cost}
This part presents visual representations to further elucidate our model's performance across different datasets. These figures are integral to understanding the effectiveness and efficiency of the proposed Active Learning framework in handling varied text classification tasks.

\begin{figure*}[t]

   \centering
    \captionsetup{justification=raggedright,singlelinecheck=false}
   \includegraphics[width=\textwidth]{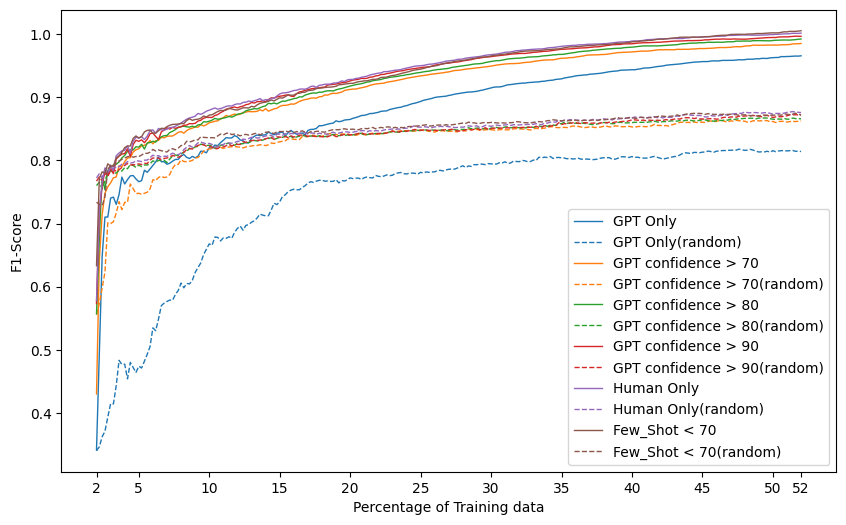}
   \caption{\label{fig:IMDB}
F1 Score Progression from 2\% to 52\% Training Data Portions in the IMDB Dataset. This figure visualizes the evolution of F1 scores across different training data portions for various annotation methods, including GPT-only, Hybrid models, Human-only, Few-shot learning, and a baseline of random sampling. }
\end{figure*}

Figure \ref{fig:IMDB} presents a detailed comparison of the model's performance on the IMDB dataset across various methods, including the proposed Active Learning framework and a baseline method of random data addition. The figure illustrates the F1 scores as a percentage for each method, showcasing the effectiveness of selective data addition in enhancing model accuracy and precision over iterative learning processes.

\begin{figure*}[t]

   \centering
    \captionsetup{justification=raggedright,singlelinecheck=false}
   \includegraphics[width=\textwidth]{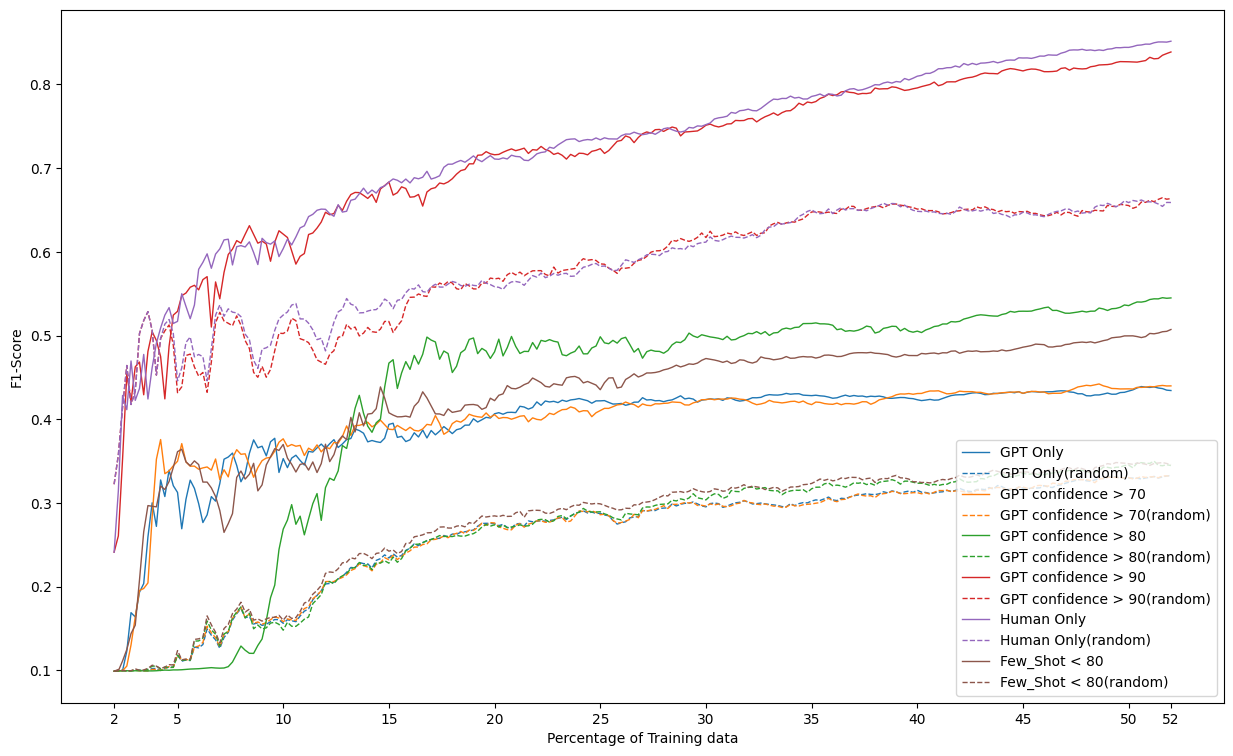}
   \caption{\label{fig:Genre}
This figure shows the visualization and evolution of F1 scores of the Movie Genres Dataset across different training data portions for various annotation methods compared with a baseline of random sampling. }
\end{figure*}

Similarly, Figure \ref{fig:Genre} provides a comparative analysis for the Movie Genres dataset. It juxtaposes the results of the Active Learning framework against the baseline random data addition method. This comparison is crucial in demonstrating the model's capability to handle different classification tasks and the superiority of the selective addition approach in improving performance metrics.

\subsection{Proxy Validation Correlation Examples}
This section of the appendix illustrates through examples the effectiveness of using a small subset of the data (5\% of the entire dataset) as a proxy validation tool. These figures demonstrate how the proxy validation F1 scores serve as reliable indicators of the overall pool quality, which is particularly important in real-world scenarios where access to the complete pool labels is not feasible.

\begin{figure*}[t]

   \centering
    \captionsetup{justification=raggedright,singlelinecheck=false}
   \includegraphics[width=\textwidth]{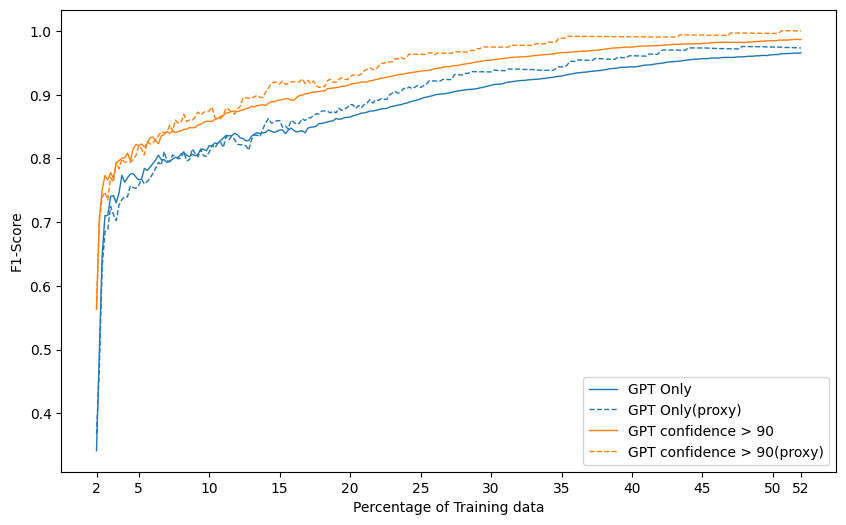}
   \caption{\label{fig:IMDB_proxy}
Correlation Between GPT-3.5 Annotation Confidence and Proxy Validation in IMDB Dataset: This figure illustrates the alignment of proxy validation F1 scores with GPT-3.5 annotations at confidence levels above 90\% and GPT-3.5-only annotation.  }
\end{figure*}

Figure \ref{fig:IMDB_proxy} showcases the correlation for the IMDB dataset. It compares the F1 scores obtained from GPT-3.5 annotations (with confidence > 90\%) and GPT-3.5 only annotation to the F1 scores from the proxy validation set. The key observation here is the trend alignment between the proxy validation scores and the overall pool quality. This figure serves as an empirical example to demonstrate how effectively the proxy validation set can estimate the model's performance and help in determining an optimal stopping point for the Active Learning process.

\begin{figure*}[t]

   \centering
    \captionsetup{justification=raggedright,singlelinecheck=false}
   \includegraphics[width=\textwidth]{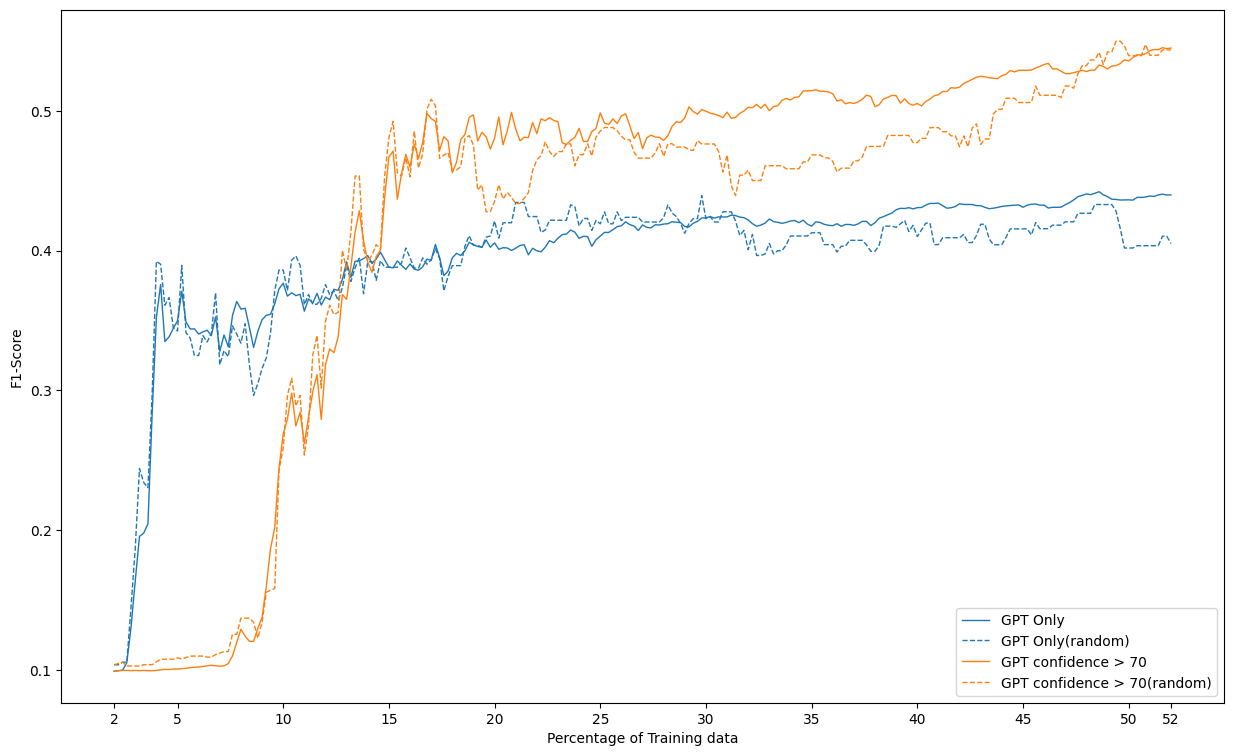}
   \caption{\label{fig:Genre_proxy}
Proxy Validation F1 Score Trends for Different GPT-3.5 Confidence Levels in Movie Genre Dataset: Displaying the trend similarity in proxy validation F1 scores for GPT-3.5 annotations with confidence levels above 70\% and GPT3.5 only.  }
\end{figure*}

Figure \ref{fig:Genre_proxy} focuses on the Movie Genres dataset and presents a similar analysis. It contrasts the F1 scores for GPT-3.5 annotations with confidence levels > 70\% and GPT-3.5 only against the proxy validation F1 scores. Despite the varying confidence levels, both scenarios consistently correlate with the proxy validation scores. This figure highlights the effectiveness of the proxy validation in mirroring the potential performance of the model in the broader dataset, thus serving as a guide for the continuation or cessation of data annotation efforts.
\subsection{Prompt Design}
To illustrate the type of prompt used for GPT-3.5 annotation, consider the example shown in Figure \ref{fig:gpt_prompt} for the IMDB dataset.

\begin{figure}[htbp]
\centering
\begin{lstlisting}[language=Python, label=lst:gpt_prompt, basicstyle=\small\ttfamily, showstringspaces=false, breaklines=true, frame=single]
prompt = f"""
    What is the sentiment of the following movie review,
    and how confident you are about this 'sentiment',
    which is delimited with triple backticks?

    Give your answer as a single word, either 'positive'
    or 'negative' and a single percentage in JSON format delimited with space.

    Review text: '''{X[i]}'''
    """
\end{lstlisting}
\caption{Example of a GPT-3.5 prompt for sentiment analysis in movie reviews, formatted to request output in JSON format.}
\label{fig:gpt_prompt}
\end{figure}

This prompt was designed to be straightforward, directing GPT-3.5 to classify a given movie review as either Positive or Negative. The simplicity of the prompt ensures clarity in the task, allowing GPT-3.5 to focus solely on the sentiment analysis of the provided review text.

The example provided in Figure \ref{fig:few_shot_genre} shows a few-shot prompt used in the study for the Movie Genres dataset.

\begin{figure}[htbp]
\centering
\begin{lstlisting}[language=Python, label=lst:few_shot_genre, basicstyle=\small\ttfamily, showstringspaces=false, breaklines=true, frame=single]
prompt = f"""
  Determine the genre of the movie based on the following plot:

  For the plot provided, classify its genre as a single word (without other marks or words like 'genre:'), either "comedy", "action", "drama", or "horror".

  Use the few-shot learning examples below to improve your prediction:

  Few-shot Examples:
  ```{newsetX.iloc[0]}``` genre:{newsetY.iloc[0]}
  ```{newsetX.iloc[1]}``` genre:{newsetY.iloc[1]}
  ```{newsetX.iloc[2]}``` genre:{newsetY.iloc[2]}
  Movie plot:
  ```{newsetX.iloc[i]}```
  """
\end{lstlisting}
\caption{Example of a GPT-3.5 few-shot learning prompt for Movie Genres classification.}
\label{fig:few_shot_genre}
\end{figure}

This prompt provides GPT-3.5 with a few examples to illustrate the task, followed by a new description for classification. The structure of the prompt is key in 'teaching' the model and the nature of the task, using just a few examples, enabling it to apply this understanding to new, unseen descriptions.

\subsection{Dataset and Active Learning details}

\begin{figure*}[t] 

   \centering
   \captionsetup{justification=raggedright,singlelinecheck=false}
   \includegraphics[width=\textwidth]{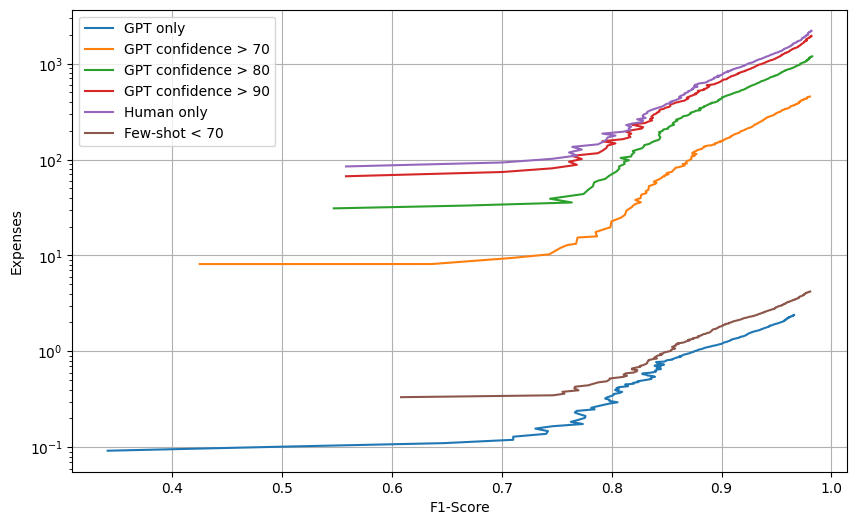}

   \caption{\label{fig:IMDB_cost}
Cost per F1 Score Analysis During Iterative Training Data Increments in the IMDB Dataset. This figure illustrates the cost efficiency (cost per F1 score) for different annotation strategies as the training data portion increases from 2\% to 52\%. It compares the cost-effectiveness of GPT-only, Hybrid models, Human-only, and Few-shot learning methods.}
\end{figure*}

\begin{figure*}[t] 

   \centering
   \captionsetup{justification=raggedright,singlelinecheck=false}
   \includegraphics[width=\textwidth]{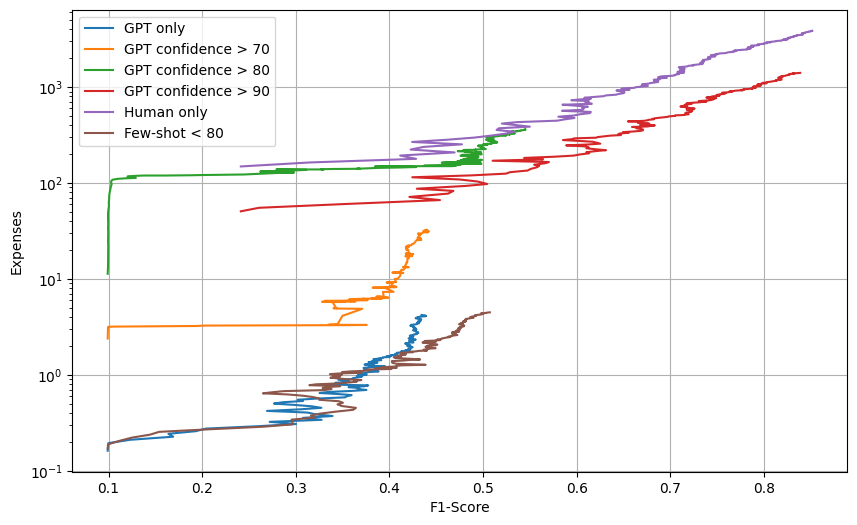}

   \caption{\label{fig:Genre_cost}
Cost per F1 Score Analysis During Iterative Training Data Increments in the Movie Genress Dataset. This figure shows the cost efficiency (cost per F1 score) for different annotation strategies as the training data portion. .}
\end{figure*}

Figure \ref{fig:IMDB_cost} and Figure \ref{fig:Genre_cost} delve into the cost implications associated with achieving different F1 scores in the IMDB dataset Movie Genres dataset, respectively. These figures provide a detailed breakdown of the costs incurred in each experiment, offering insights into the economic feasibility and efficiency of models in achieving high levels of accuracy at a reduced cost.


The sizes of the datasets used in our research were carefully chosen to ensure a comprehensive analysis while maintaining manageability. The IMDB dataset consisted of 10,000 entries, providing a rich source of movie reviews for sentiment analysis. The fake news dataset comprised 5,000 entries, offering a diverse range of articles for the identification of veracity in news content. Lastly, the Movie Genres dataset included 4,000 entries, encompassing various movie descriptions for genre classification. A critical aspect of our dataset selection was the balance in each dataset. We ensured that each dataset was carefully balanced to represent a wide range of scenarios and conditions. This balance was crucial in avoiding biases and ensuring that the results of our study were fair and unbiased.

The Active Learning phase of our study began with an initial dataset comprising 2\% of the total data for each dataset. This initial selection served as the baseline for our model. From this point, we engaged in a systematic and incremental learning process, expanding the dataset by 0.002\% in each iteration. This process was repeated over a total of 250 iterations. By the end of these iterations, we had cumulatively added an additional 50\% of data to our initial pool, bringing the total data utilized to 52\%. This gradual and iterative approach was critical in optimizing the learning curve of the model, allowing it to progressively improve its classification accuracy while being exposed to more data samples.

\end{document}